\documentclass[runningheads]{llncs}

\usepackage{cite}
\usepackage{amsmath,amssymb,amsfonts}
\usepackage{algorithmic}
\usepackage{graphicx}
\usepackage{textcomp}
\usepackage{xcolor}
\usepackage{url}
\usepackage{amsmath}
\usepackage{amssymb}
\usepackage{bbm}
\usepackage{dsfont}
\usepackage[linesnumbered,ruled,vlined]{algorithm2e}
\SetKwInput{KwInput}{Input}     
\SetKwInput{KwOutput}{Output}     
\usepackage{mathtools}
\usepackage{multirow}
\usepackage{booktabs}
\usepackage[normalem]{ulem}

\usepackage{todonotes}

\def\BibTeX{{\rm B\kern-.05em{\sc i\kern-.025em b}\kern-.08em
    T\kern-.1667em\lower.7ex\hbox{E}\kern-.125emX}}

\title{Contrastive Representations \\ for Label Noise Require Fine-Tuning}

\titlerunning{Contrastive Representations for Label Noise Require Fine-Tuning}

\author{Pierre Nodet\inst{1,}\inst{2} \and
Vincent Lemaire\inst{1} \and\\
Alexis Bondu\inst{1} \and
Antoine Cornuéjols\inst{2}}

\authorrunning{P. Nodet et al.}

\institute{Orange Labs, Paris \& Lannion, France
\and
AgroParisTech, Paris, France\\
}

\begin{document}

\maketitle

\begin{abstract}
In this paper we show that the combination of a Contrastive representation with a label noise-robust classification head requires fine-tuning the representation in order to achieve state-of-the-art performances. Since fine-tuned representations are shown to outperform frozen ones, one can conclude that noise-robust classification heads are indeed able to promote meaningful representations if provided with a suitable starting point. Experiments are conducted to draw a comprehensive picture of performances by featuring six methods and nine noise instances of three different kinds (none, symmetric, and asymmetric). In presence of noise the experiments show that fine tuning of Contrastive representation allows the six methods to achieve better results than end-to-end learning and represent a new reference compare to the recent state of art. Results are also remarkable stable versus the noise level.
\end{abstract}

\section{Introduction}

Deep Learning (DL) paradigm has proved very powerful in many tasks, however recent papers \cite{DBLP:conf/nips/MaennelATBBGK20,zhang2021understanding} have shown that \textbf{``noisy labels"} are a real challenge for end-to-end deep learning architectures. Their test performance is found to deteriorate significantly even if they are able to learn perfectly the train examples. This problem has attracted a lot of suggestion in many recent papers. 

Zhang et al. \cite{zhang2020decoupling} conducted experiments to analyze the impact of label noise on deep architectures, and they found that the performance degradation mainly comes from the representation learning rather than the classification part.
It therefore appears very difficult to learn a relevant representation in the presence of label noise, in an end-to-end manner.

To tackle this problem, one option is to exploit an already existing representation which has been learned in an unsupervised way. 
In particular, Self Supervised Learning \cite{jing2020self} (SSL) gathers an ensemble of algorithms which automatically generate supervised tasks from unlabeled data, and, therefore to learn representations from examples that are not affected by label noise. 
An example of SSL algorithm is Contrastive Learning \cite{jaiswal2021survey}, where a representation of the data is learned by making feature vectors from similar
pictures (i.e. generated from the same original picture by using two different transformer functions) to be close in the feature space whereas feature vectors from dissimilar pictures are to be far apart.
In \cite{ghosh2021contrastive}, the authors propose to initialize the representation with a pre-trained Contrastive Learning one, and then, to use the noisy labels to learn the classification part and fine-tune the representation. 
It appears that this approach clearly outperforms the end-to-end architecture, where the representation is learned from noisy labels.

But questions remain: is this performance improvement only attributable to the quality of the Contrastive Representation used (i.e. the starting point of fine-tuning)? Or is the fine-tuning step able to promote a better representation? To answer these questions this paper examines the different possibilities to learn a DL architecture in presence of label noise: (i) end-to-end learning (ii) learning only the head part when freezing a contrastive representation and (iii) fine tuning the later representation.

The rest of this paper is organized as follows. The section 2 provides a brief overview of the main families of algorithms dedicated to fight the label noise underlying the issue of preserving a good representation in spite of label noise.  Section 3 then describes the experimental protocol. The section 4 will present the results and a deep analysis which will allow us to answer the questions above. The last section raises an interesting conclusion and provides some perspectives for future work.

\section{Representation Preserving with Noisy Labels}
\label{sec_preserving}

This section presents a brief overview of the state of the art on \textit{learning deep architecture with noisy labels} emphasizing how these methods \textit{preserve}, to some extent, \textit{the learned representation} in the presence of label noise.  For an extended overview, the reader may look \cite{song2021learning}.

\subsection{Preserving by Recovering}
\label{recovering}

The dominant approach to preserve the learned representation is to recover a clean distribution of the data from the noisy dataset. It mostly consists in finding a mapping function from the noisy to the clean distribution thanks to heuristics, algorithms or machine learning models. Three different ways of recovering the clean distribution are usually put forward \cite{nodet2020importance}: (i) sample reweighting; (ii) label correction and (iii) instance moving.

\smallskip
\noindent
\textbf{Recovering by Reweighting} - 
The sample reweighting methodology aims at assigning a weight to every samples such that the reweighted population behaves as being sampled from the clean distribution. 
The Radon-Nikodym derivative (RND) \cite{Nikodym1930} of the clean concept with respect to the noisy concept is the function that defines the perfect reweighting scheme. Many algorithms therefore rely on providing a good estimation of the RND by learning it from the data using Meta Learning \cite{pmlr-v80-ren18a} or minimizing the Maximum Mean Discrepancy of both distributions in a Reproducing kernel Hilbert space \cite{Liu_2016,fang2020rethinking}. Many of these methods are inspired by the covariate shift problem 
\cite{huang2007correcting,gretton2009covariate}.
Other algorithms rely on different reweighting schemes that do not involve the RND as done, for instance, in Curriculum Learning \cite{bengio2009curriculum}. They are described in details later in this section.
By doing sample reweighting, algorithms evaluate whether or not a sample is deemed to have been corrupted and assign a lower weight to a suspect sample so that its influence on the training procedure is lowered. The hope is that clean samples are sufficient to learn high-quality representations

\smallskip
\noindent
\textbf{Recovering by Relabelling} -
Another way to recover the clean distribution from the noisy data is to correct the noisy labels. 
One great advantage over sample reweighting is that corrected samples can be fully used during the training procedure. Indeed, when a sample is corrected, it will count as one entire sample in the training procedure (gradient descent for example), whereas a reweighted noisy sample would get a low weight and would not be used significantly in the training procedure. Thus, when done effectively, label correcting might get better performance. Meta Label Correction (MLC) \cite{shu2020meta} is an example of this approach where the label correction is done thanks to a model learned using meta learning. %
One downside of label correction, however, is that the label of a clean sample can get ``corrected'' or the label of a noisy sample can get changed to a wrong label. 
Label correcting algorithm assign the same weight to all training examples, even though they might have ``corrected'' a label based on shaky assumptions.
By contrast, Sample reweighting will assign a low weight if the algorithm is not confident in whether the sample is clean or noisy. 

\smallskip
\noindent
\textbf{Recovering by Modifying} -
A third way to recover the clean distribution is by modifying the sample itself so that its position in the feature space gets closer or is moved within an area for its label that seems more appropriate (i.e. obeying regularisation criteria). Finding a transformation in the latent space itself has the advantage to require less labelled samples, or even none at all, as the work is performed on distance between samples themselves, like for example in  \cite{shi2012information}.

\subsection{Preserving by Collaboration}
\label{collaboration}

Multiple algorithms and agreements measures have been used in many sub-fields of machine learning such as ensembling \cite{breiman_bagging_1996,freund1997decision,friedman2001} or semi supervised learning \cite{yarowsky1995unsupervised,blum1998combining}. They can be adapted to learn with noisy labels by relying on a disagreement method between models in order to detect noisy samples. When the learned models disagree on predictions for the label of a sample, this is considered as a sign that the label of this sample may be noisy. When the models used are diverse enough, these methods are often found to be quite efficient \cite{han_CoteachingRobustTraining_2018, wang2020collaborative}. 

However these algorithms suffer from learning their own biases and diversity needs to be introduced in the learning procedure. Using algorithms from different classes of models and different origins can increase the diversity among them by introducing more source of biases \cite{Lee2020Robust}. Alternating between learning from the data and from the other models is another way to combat the reinforcement of the models' biases \cite{wang2020collaborative}. These algorithms rely on carefully made heuristics to be efficient.

\subsection{Preserving by Correcting}
\label{correcting}

When learning loss base models, such as neural networks, on label noise, the loss value of a training example can be a discriminative feature to decide if its label is noisy. Deep neural networks seem to have the property that they first learn general and high level patterns from the data before falling prey to overfitting the training samples, especially in the presence of noisy labels \cite{pmlr-v70-arpit17a,pmlr-v108-li20j}. As they are ``learned'' at a later stage, these noisy examples are often associated with a high loss value \cite{tocheck} which may then highly influences the training procedure and perturb the learned representation \cite{zhang2021understanding}. A way to combat label noise is accordingly to focus first on small loss and easy examples and keep the high loss and hard examples for the end of the training procedure. Curriculum Learning \cite{bengio2009curriculum} is a way to employ this training schema with heuristic based schemes \cite{NIPS2010_e57c6b95,felzenszwalb2008discriminatively,jiang2015self,lin2017focal} or schemes learned from data \cite{pmlr-v80-jiang18c,shu2019mwnet}. This class of algorithm has the same properties as the ones relying on importance reweighting, but maybe more adapted to training with iterative loss based algorithms such as neural networks or linear models.

Instead of filtering or reweighting samples based on their loss values, one could try to correct the loss for these samples using the underlying noise pattern. Numerous method have been doing so by estimating the noise transition matrix for \textit{Completely at Random} (i.e uniform) and \textit{At Random} (i.e class dependent) noise \cite{patrini2017making,Hendrycks2018,shu2020meta}. This category of algorithms are still to be tested on more complex noises scenarios such as \textit{Not at Random} (i.e instance and class dependent) noise.

\subsection{Preserving by Robustness}
\label{robust}

The last identified way to preserve the learned representation of a deep neural network in presence of label noise is by using a robust or regularized training procedure. This can take multiple forms from losses to architectures or even optimizers. One of them are Symmetric Losses \cite{NIPS2015_5941,Ghosh_Kumar_Sastry_2017,charoenphakdee_symmetric_2019}. A symmetric loss has the property that: $\forall x \in \mathcal{X}$, $\sum_{y \in \mathcal{Y}} L(f(x), y)=c$ where $c \in \mathbb{R}$. These losses have been proven to be theoretically insensitive to Completely at Random (CAR) label noise. 
Recently, modified versions of the well-known Categorical Cross Entropy (CCE) loss have been designed in order to be more robust and thus more resistant to CAR label noise as is the case for the Symmetric Cross Entropy (SCE) loss \cite{wang2019symmetric} or the Generalized Cross Entropy Loss (GCE) \cite{NEURIPS2018_f2925f97}. Both of these rely on using the CCE loss combined with a  known more robust loss such as the Mean Absolute Error (MAE). 
However, the resulting algorithms often underfit in presence of too few label noise while they are unable to learn a correct classifier with too much label noise.

All these approaches still adopt the end-to-end learning framework, aiming at fighting the effects of label noise by preserving the learned representation. However they fail to do so in practice: \textit{decoupling} the learning of the representation, using Self Supervised (SSL) learning, from the classification learning stage itself and then fine tuning the representation with robust algorithms is beneficial for the model performance \cite{zhang2020decoupling,ghosh2021contrastive}. A natural question arises about the origin of the performance improvements, and the ability of these algorithms to learn or promote a good representation in presence of label noise. If robust algorithms are unable to learn a representation it should be even better to freeze the SSL representation instead of fine tuning it.

In order to assess the origin of the improvements for different classes of algorithms and different noise levels, we compare the above-mentioned end-to-end approaches against each other when the representation is learned in a self-supervised fashion by either fine tuning or freezing the representation when the classification head is learned. Thus, any difference in the performance would be attributable to the difference in the representation learnt.

\section{Experimental Protocol}
\label{experiments}

In \cite{zhang2020decoupling}, the authors showed that when using end-to-end learning, fine tuning the representation on noisy labels harms a lot the final performance, while learning a classifier on frozen embeddings is quite robust to label noise and leads to significant performance improvements over state-of-the-art algorithms if the representation is learned using trustful examples. The latter can be found for instance using confidence and loss value. 
Nonetheless it is arguable whether these improvements were brought by an efficient self-supervised pretraining (SSL) with SimCLR \cite{chen2020simple}, a contrastive learning method, or by the classification stage of the REED algorithm \cite{zhang2020decoupling}.

The goal of the following experimental protocol is to assess and isolate the role of the contrastive learning stage, in the performance that can be achieved by representative methods as presented in Section \ref{sec_preserving} about state of the art approaches. Specifically, several RLL algorithms have been chosen, one from each of the highlighted families (see Section \ref{sec_preserving} and Table \ref{knowledge}). For each, the difference in performance between using contrastive learning to learn the representation and the performance reported with the original end-to-end algorithms is measured. These experiments seek to highlight the impact of each RLL algorithms and assess if these are able to promote a better representation than the pretrained contrastive representation through fine-tuning.

The rest of this section describes the experimental protocol used to conduct this set of experiments.

\subsection{The tested Algorithms}

Section \ref{sec_preserving} presented an overview of the state of the art for learning with label noise organized around families of approaches that we highlighted. Since our experiments aim at studying the properties of each of these approaches, we selected one representative technique from each of these families as indicated in the following.  

\begin{itemize} 
\item In the first family of techniques (\textit{recover the clean distribution}), the algorithms re-weight the noisy examples or attempt to correct their label.     One of these algorithm uses what is called Dynamic Importance Reweigthting {\bf (DIW)}. It reweights samples using Kernel Mean Matching (KMM) \cite{huang2007correcting,gretton2009covariate} as is done in covariate shift with Density Ratio Estimators \cite{sugiyama2010density}. Because this algorithm adapts well-grounded principles to end-to-end deep learning, it is a particularly relevant algorithm for our experiments.
    
\medskip    
\item Co\-Learning {\bf(CoL)} \cite{wang2020collaborative} is a good representative of the family of \textit{collaborative learning algorithms}. It uses disagreements criteria to detect noisy labels and is tailored for end-to-end deep learning where the two models are branches of a larger neural networks. It appears to be one of the best performing collaborative algorithm while not resorting to complex methods such as data augmentation or probabilistic modelling like the better known DivideMix \cite{li2019dividemix}.
    
\medskip
\item The third identified way to combat label noise is by \textit{mitigating the effect of high loss samples} \cite{tocheck} by either ditching them or using a loss correction approach. Curriculum learning is often used to remove the examples that are associated with high loss from the training set. {\bf (MWNet)} \cite{shu2019mwnet} is one the most recent approach using this technique, which learns the curriculum from the data with meta learning. Besides, Forward Loss Correction {\bf (F-Correction)} \cite{patrini2017making} and Gold Loss Correction {\bf(GLC)} \cite{Hendrycks2018} are two of the most popular approaches to combat label noise by \textit{correcting the loss function}. Both seek to estimate the transition matrix between the noisy labels to the clean labels, the first technique using a supervised approach thanks to a clean validation set, and the second one in an unsupervised manner. Even though many extensions of these algorithm have been developed since then \cite{Xia2019AreAP,shu2020meta}, in these experiments, we use F-Correction and GLC since they are way simpler and almost as effective.

\medskip
\item Lastly, in recent literature, a new emphasis is put on the research of new loss functions that are conducive to better risk minimization in presence of noisy labels \textit{for robustness purpose}. For example, \cite{NIPS2015_5941,charoenphakdee_symmetric_2019} show theoretically and experimentally that when the loss function satisfies a symmetry condition, described below, this contributes to the robustness of the classifier. The Generalized Cross Entropy {\bf (GCE)} \cite{NEURIPS2018_f2925f97} is the robust loss chosen in this benchmark as it appears to be very effective.

\end{itemize}

\begin{table}[t]
\begin{center}
\begin{tabular}{@{}lccl@{}}
\toprule
\textbf{Algorithms (Date)}  & \textbf{Noise Ratio} & \textbf{Clean Validation} & \textbf{Family (Section)} \\
\midrule
DIW (2020) & $\times$ & $\checkmark$ & Reweighting (\ref{recovering})\\
CoLearning (2020) & $\checkmark$ & $\times$ & Collaborative Learning (\ref{collaboration})\\
MWNet (2019) & $\times$ & $\checkmark$ & Curriculum Learning (\ref{correcting})\\
F-Correction (2017) & $\times$ & $\times$ & Loss Correction (\ref{correcting})\\
GLC (2018) & $\times$ & $\checkmark$ & Loss Correction (\ref{correcting})\\
GCE (2018) & $\times$ & $\times$ & Robust Loss (\ref{robust})\\ 
\bottomrule
\end{tabular}
\end{center}
\caption{Taxonomy of robust deep learning algorithms studied in this paper. The \textbf{Noise Ratio} column corresponds to whether the algorithm needs the noise rate ($\checkmark$) to learn from noisy data or not ($\times$). The \textbf{Clean Validation} column corresponds to whether the algorithm needs an additional clean validation dataset ($\checkmark$) to learn from noisy data or not ($\times$).}
\label{knowledge}
\end{table}

A note \textit{about additional requirements}: These algorithms may have additional requirements, mostly some knowledge about the noise properties. These are described in table \ref{knowledge}. In the experiments presented below, the clean validation dataset is set to be 2 percent of the total training data, like in \cite{shu2019mwnet,zheng2021mlc},  and the noise probability is provided to the algorithms that need it.

A note \textit{about the choice of the pretrained architecture}: We chose to use SimCLR for Self-Supervised Learning (SSL) as done in \cite{zhang2020decoupling}.

\begin{figure}[h]
\begin{center}
\includegraphics[width=0.5\linewidth]{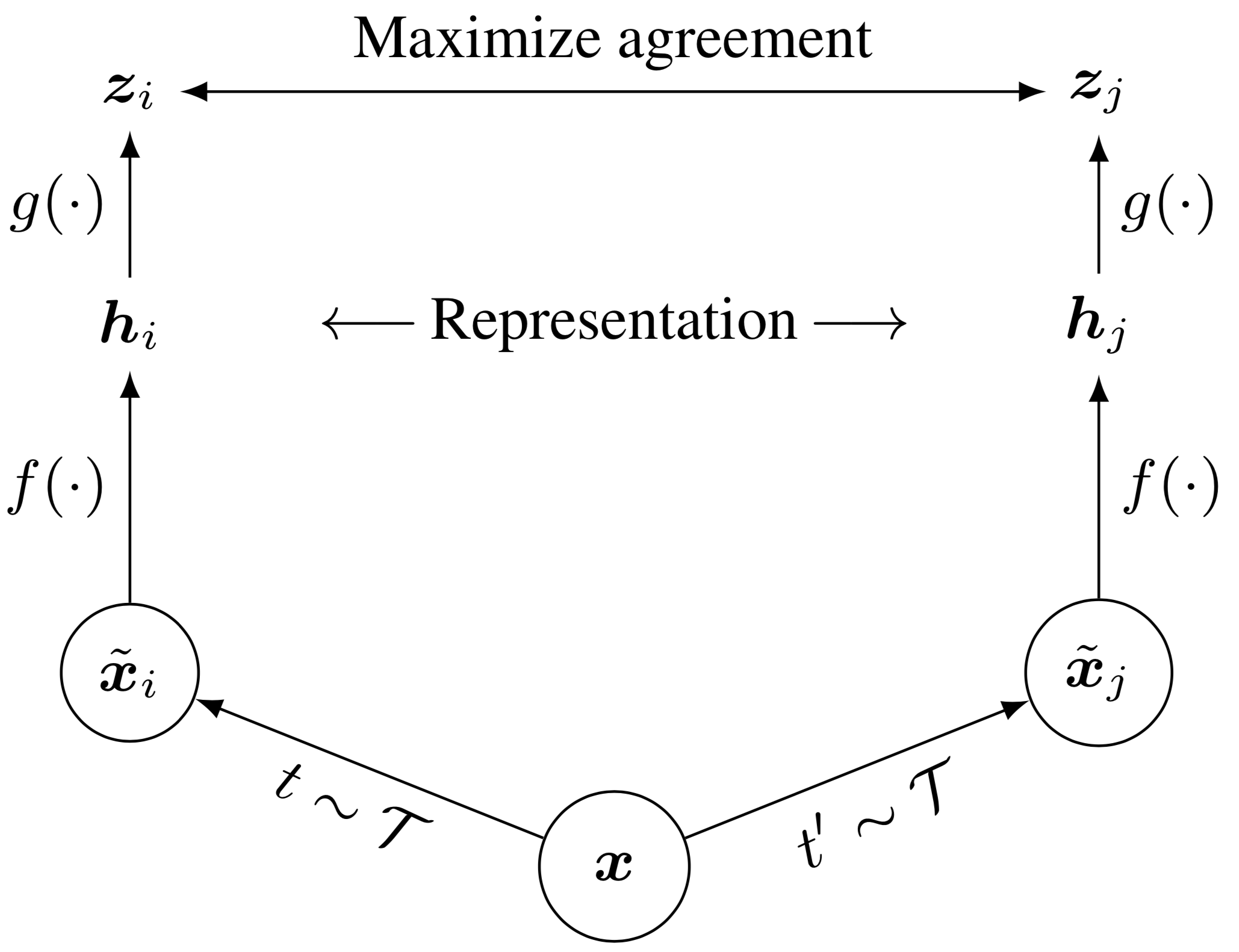}
\end{center}
\caption{Figure from \cite{chen2020simple}: ``A simple framework for contrastive learning of visual
representations. Two separate data augmentation operators are
sampled from the same family of augmentations ($t \sim \mathcal{T}$ and
$t' \sim \mathcal{T}$ ) and applied to each data example to obtain two correlated
views. A base encoder network $f$(·) and a projection head $g$(·)
are trained to maximize agreement using a contrastive loss. After
training is completed, we throw away the projection head $g$(·) and
use encoder $f$(·) and representation $\mathbf{h}$ for downstream tasks." }
\label{fig:simclr}
\end{figure}

SimCLR is a contrastive learning algorithm that is composed of three main components (See Figure \ref{fig:simclr}): a family of data augmentation $\mathcal{T}$, an encoder network $f$(·) and a projection head $g$(·). Data augmentation is used as a mean to generate positive pairs of samples: a single image $\mathbf{x}$ is transformed into two similar images $\tilde{\mathbf{x}}_i$ and $\tilde{\mathbf{x}}_j$ by using a data augmentation module $\mathcal{T}$ with different seeds $t$ and $t'$. Then the two images go through an encoder network $f$(·) to extract an image representation $\mathbf{h}$, such as $\mathbf{h}_i=f(\tilde{\mathbf{x}}_i)$ and $\mathbf{h}_j=f(\tilde{\mathbf{x}}_j)$. Finally a projection  head $g$(.) is used to train the contrastive objective in a smaller sample space $\mathbf{z}$, with $\mathbf{z}_i=g(\tilde{\mathbf{h}}_i)$ and $\mathbf{z}_j=g(\tilde{\mathbf{h}}_j)$. The contrastive loss used is called the NT-Xent, the normalized temperature-scaled cross entropy loss, and defined by the following formula:

\begin{equation}
\ell(\mathbf{z}_i,\mathbf{z}_j) = -\log\frac{\exp(\text{sim}(\mathbf{z}_i, \mathbf{z}_j )/\tau )}{
\sum_{k=1}^{2N} \exp(\text{sim}(\mathbf{z}_i, \mathbf{z}_k)/\tau )}
\end{equation}

where $\tau$ is the temperature scaling and $\text{sim}$ is the cosine similarity. The final loss is computed across all positive pairs, both $(i, j)$ and $(j, i)$, in a mini-batch. When the training of SimCLR is complete, the projection head $g$(.) is dropped and the embeddings $\mathbf{h}$ are used as an image representation in downstream tasks.

Other SSL algorithms could have been used as well, such as Moco \cite{he2019moco,chen2020mocov2} or Bootstrap Your Own Latent (BYOL) \cite{NEURIPS2020_f3ada80d}. However, we do not expect that the main conclusions of the study would be much changed.

\subsection{Datasets}
\label{datasets}

The datasets chosen in this benchmark are two image classification datasets namely CIFAR10, CIFAR100. They are two famous image classification datasets, containing only clean examples and as such, we will simulate symmetric (Completly at Random) and asymmetric (At Random) noise as defined later in section \ref{simulated}. These benchmarks should be extended to other image classification datasets such as FashionMNIST, Food-101N, Clothing1M and Webvision and to other classification tasks such as text classification or time series classification.

\subsection{Simulated Noise}
\label{simulated}

As datasets chosen in Section \ref{datasets} contains clean labels, label noise will be introduced synthetically on the training samples. Two artificial noise models will be used, a symmetric (Completely at Random) and asymmetric (At Random) noise. Symmetric noise corrupts a label from one class to any other classes with the same probability, meanwhile the asymmetric corrupts a label to a similar class only. Similar classes are defined through class mappings. For CIFAR-10, the class mappings are TRUCK $\rightarrow$ AUTOMOBILE, BIRD$\rightarrow$ AIRPLANE, DEER $\rightarrow$ HORSE, CAT $\leftrightarrow$ DOG. For CIFAR-100, the class mappings are generated from the next class in that group (where 100 classes are categorized into 20 super-classes of 5 classes). These class mappings are the ones introduced in \cite{patrini2017making, NEURIPS2018_f2925f97}.

\subsection{Implementation Details}
\label{implementationdetails}

We give some implementation details for reproducibility and / or a better understanding of the freezing process in the experiments:

\begin{itemize}
\item On CIFAR10 and CIFAR100 the SGD optimizer will be used to train the final Multinomial Logistic Regression with an initial learning rate of $0.01$, a weight decay of $1e^{-4}$ and a non-Nesterov momentum of $0.9$. The learning rate will be modified during training with cosine annealing \cite{loshchilov2016sgdr}. The batch size is 128.

\item When doing the \textit{"Freeze"} experiments, the weights of SimCLR from \cite{ghosh2021contrastive} will be used and will not be modified during the training procedure. All the weights up to before the projection head of SimCLR are used, then the dimension output of the feature encoder is 2048 for CIFAR10 and CIFAR100. The classification architecture is composed by a single linear layer with an output dimension of 10 (or 100), corresponding to the number of classes. Thus when trained with the Categorical Cross Entropy it corresponds to a usual logistic regression. This classifier is going to be learned with multiple algorithms robust to label noise. These algorithms are not modified from their original formulation.

\item The \textit{"Fine Tuning"} experiments follow the same implementation as the \textit{"Freeze"} experiments. However the weights of the same pretrained SimCLR encoder are allowed to be modified by backpropagation.

\item Based on their public implementation and / or article  we re-implemented all the algorithm tested (DIW \cite{fang2020rethinking}, CoL \cite{wang2020collaborative},
MWNet \cite{shu2019mwnet}, F-Correction \cite{patrini2017making}, GLC \cite{Hendrycks2018} and GCE \cite{NEURIPS2018_f2925f97}). All these re-implemented algorithms will soon be available as an open source library easily usable by researchers and practitioners. These custom implementations have been verified to produce, under the same condition stated in the corresponding original papers (noise models, network architectures, optimizers, ...), the same results or results in the interval of confidence (for clean or noisy labels). We may thus be confident that results in the different parts of the Tables \ref{table-results-cifar10} and \ref{table-results-cifar100} are comparable.

\item The experiments have been run multiple times for all algorithms, some datasets, some noise models and some noise ratios with different seeds to see the seed impact on the final performance of the classifier. For all algorithms, the standard deviation of the accuracy was less than 0.1 percent.

\end{itemize}

\section{Results}
\label{sec_results}

This section reports the results obtained using the protocol described in section \ref{experiments}. They are presented in the tables \ref{table-results-cifar10} and \ref{table-results-cifar100} corresponding to the two tested datasets CIFAR10 and CIFAR100. Each table is composed of four rows subsections corresponding to the different types of representation used, which can be learned in a \textit{End-to-End} manner (A), be taken from an already existing SSL model, either \textit{Frozen} (B) or \textit{Fine tuned} (C).
Moreover they are composed of two columns subsections corresponding to the noise model used to corrupt samples (symmetric or asymmetric).

These tables present the results from different studies: (A) The first part of these tables about {\it ``End-to-End learning''} are results reported in the respective papers \cite{fang2020rethinking,wang2020collaborative,shu2019mwnet,patrini2017making,Hendrycks2018,NEURIPS2018_f2925f97} or reported in \cite{ghosh2021contrastive}; (B) The second part about {\it ``Freeze''} experiments conducted in this paper, are made by re-implementing the referred algorithms from scratch; (C) The {\it``Fine Tuning''} experiments are results reported in \cite{ghosh2021contrastive}.

\begin{table}[t]
\fontsize{8}{9}\selectfont
\begin{center}
\begin{tabular}{l|l|ccccccccc}
\toprule
\multicolumn{2}{c|}{\multirow{3}{*}{Algorithms}} & \multicolumn{9}{c}{CIFAR10} \\
\multicolumn{2}{c|}{} & Clean & \multicolumn{6}{c}{Symmetric} & \multicolumn{2}{c}{Asymmetric} \\
\multicolumn{2}{c|}{} & 0 & 20 & 40 & 60 & 80 & 90 & 95 & 20 & 40\\
\midrule
DIW \cite{fang2020rethinking} & \multirow{6}{*}{$\text{End-to-End}$ (A)} &  &  & 80.4 & 76.3 &  &  &  & & 84.4 \\
CoL \cite{wang2020collaborative} &  &  & 93.3 & 91.2 &  & 49.2 &  &  & 88.2 & 82.9  \\
MWNet \cite{shu2019mwnet} &  & 95.6 & 92.4 & 89.3 & 84.1 & 69.6 & 25.8 & 18.5 & 93.1 & 89.7 \\
F-Correction \cite{patrini2017making} &  & 90.5 & 87.9 &  &  & 63.3  & 42.9 &  & 90.1 & \\
GLC \cite{Hendrycks2018} &  & 95.0 & 95.0 & 95.0 & 95.0 & 90.0 & 80.0 & 76.0\\
GCE \cite{NEURIPS2018_f2925f97} &  & 93.3 & 89.8 & 87.1 & 82.5 & 64.1 &  &  & 89.3 & 76.7  \\
\midrule
DIW & \multirow{6}{*}{$\text{Freeze}$ (B)} & 91.3 & 91.2 & 90.8 & 90.5 & 89.8 & 89.2 & 88.1 & 91.0 & 90.6\\
CoL & & 91.1 & 91.1 & 90.9 & 90.6 & 89.9 & 89.4 & 88.8 & 90.8 & 89.9\\
MWNet & & 91.3 & 91.2 & 90.8 & 90.6 & 89.8 & 88.2 & 82.4 & 90.9 & 86.4\\
F-Correction & & 90.8 & 90.5 & 90.1 & 89.6 & 88.4 & 88.0 & 88.1 & 88.9 & 88.4\\
GLC &  & 90.7 & 89.7 & 90.0 & 89.5 & 89.0 & 88.5 & 88.3 & 88.7 & 88.2\\
GCE &  & 91.1 & 90.8 & 90.7 & 90.5 & 90.4 & 90.0 & 89.1 & 90.9 & 89.0 \\
\midrule
DIW & \multirow{6}{*}{$\text{Fine Tuning}$ (C)} & 94.5 & 94.5 & 94.5 & 94.5 & 94.0 & 92.0 & 89.1 & 94.2 & 93.6\\
CoL & & 93.9 & 94.6 & 94.6 & 94.2 & 93.6 & 92.7 & 91.7 & 94.0 & 93.7 \\
MWNet \cite{ghosh2021contrastive} & & 94.6 & 93.9 & \multicolumn{2}{c}{92.9} & 91.5 & 90.2 & 87.2 & 93.7 & 92.6 \\
F-Correction & & 94.0 & 93.4 & 93.1 & 92.9 & 92.3 & 91.4 & 90.0 & 93.6 & 92.8\\
GLC & & 93.5 & 93.4 & 93.5 & 93.1 & 92.0 & 91.2 & 88.3 & 93.2 & 92.1 \\
GCE  \cite{ghosh2021contrastive} & & 94.6 & 94.0 & \multicolumn{2}{c}{92.9} & 90.8 & 88.4 & 83.8 & 93.5 & 90.3 \\
\bottomrule
\end{tabular}
\end{center}
\caption{Final accuracy for the different models on CIFAR10 under symmetric and asymmetric noises and multiple noise rates. }
\label{table-results-cifar10}
\end{table}

\begin{table}[t]
\fontsize{8}{9}\selectfont
\begin{center}
\begin{tabular}{l|l|ccccccccc}
\toprule
\multicolumn{2}{c|}{\multirow{3}{*}{Algorithms}} & \multicolumn{9}{c}{CIFAR100} \\
\multicolumn{2}{c|}{} & Clean & \multicolumn{6}{c}{Symmetric} & \multicolumn{2}{c}{Asymmetric} \\
\multicolumn{2}{c|}{} & 0 & 20 & 40 & 60 & 80 & 90 & 95 & 20 & 40 \\
\midrule
DIW \cite{fang2020rethinking} & \multirow{6}{*}{$\text{End-to-End}$ (A)} &  &  & 53.7 & 49.1 &  &  & &  & 54.0 \\
CoL \cite{wang2020collaborative} & & & 75.8 & 73.0 &  & 32.8 &  &  &  &   \\
MWNet \cite{shu2019mwnet} &  & 79.9 & 74.0 & 67.7 & 58.7 & 30.5 & 5.2 & 3.0 & 71.5 & 56.0 \\
F-Correction \cite{patrini2017making} &  & 68.1 & 58.6 &  &  & 19.9 & 10.2 &  & 64.2 & \\
GLC \cite{Hendrycks2018} &  & 75.0 & 75.0 & 75.0 & 62.0 & 44.0 & 24.0 & 12.0 & 75.0 & 75.0 \\
GCE \cite{NEURIPS2018_f2925f97} &  & 76.8 & 66.8 & 61.8 & 53.2 & 29.2 & & & 66.6 & 47.2\\
\midrule
DIW & \multirow{6}{*}{$\text{Freeze}$ (B)} & 65.6 & 65.1 & 64.0 & 62.9 & 59.0 & 53.3 & 42.5 & 61.7 & 49.0 \\
CoL & & 65.8 & 65.0 & 64.0 & 63.4 & 62.3 & 60.0 & 57.0 & 64.1 & 58.6\\
MWNet & & 66.6 & 66.6 & 66.2 & 65.4 & 63.7 & 59.8 & 49.5 & 64.8 & 54.5 \\
F-Correction & & 66.5 & 64.7 & 61.8 & 58.8 & 54.5 & 51.7 & 50.8 & 58.4 & 56.5 \\
GLC & & 58.5 & 57.8 & 52.3 & 51.1 & 41.6 & 40.1 & 35.3 & 51.4 & 50.3\\
GCE & & 63.5 & 62.9 & 61.5 & 60.0 & 55.7 & 51.0 & 49.9 & 51.2 & 48.3 \\
\midrule
DIW & \multirow{6}{*}{$\text{Fine Tuning}$ (C)} & 73.8 & 74.9 & 74.9 & 74.5 & 70.2 & 62.3 & 50.4 & 71.8 & 62.8\\
CoL & & 73.7 & 74.8 & 74.8 & 75.0 & 73.2 & 67.3 & 62.0 & 72.6 & 70.3\\
MWNet  \cite{ghosh2021contrastive}&  & 75.4 & 73.2 & \multicolumn{2}{c}{69.9} & 64.0 & 57.6 & 44.9 & 72.2 & 64.9  \\
F-Correction & & 69.8 & 70.1 & 69.1 & 69.5 & 66.9 & 62.1 & 57.0 & 70.3 & 66.2\\
GLC & & 69.7 & 69.4 & 68.6 & 62.5 & 50.4 & 32.1 & 18.7 & 68.2 & 62.3\\
GCE  \cite{ghosh2021contrastive}& & 75.4 & 73.3 & \multicolumn{2}{c}{70.1} & 63.3 & 55.9 & 45.7 & 71.3 & 59.3 \\
\bottomrule
\end{tabular}
\end{center}
\caption{Final accuracy for the different models on CIFAR100 under symmetric and asymmetric noises and multiple noise rates.}
\vspace{-6mm}
\label{table-results-cifar100}
\end{table}

The interpretation of the Table \ref{table-results-cifar10} and \ref{table-results-cifar100} will be done in two times, first a comparison between whole blocks (as (A) against (B)) will give insights on how deep neural networks learn representations on noisy data and how robust algorithms helps to improve the learning process or helps to preserve a given representation. Then in a second time comparisons in a given block will be made against multiple algorithms to see how well these conclusions works on different preservation families given in Section \ref{sec_preserving}.

First, we observe when comparing section (A) and (B) from both tables that \textit{"Freeze"} experiments consistently outperforms \textit{"End-to-End"} experiments as soon as the data stop being perfectly clean. Using a pretrained self-supervised representation such as SimCLR improves significantly the performances of the final classifier. Outside of well controlled and perfectly clean datasets all selected algorithms are not able to learn a good enough representation from the noisy data and are beaten by a representation learned without resorting to using given labels. Robust Learning to Label noise algorithms, especially designed for deep learning, can preserve an already good representation from noisy labels but are unable to learn a good representation from scratch.

Then, we observe when comparing section (B) and (C) from both tables that \textit{"Fine Tuning"} experiments consistently outperforms  \textit{"Freeze"} at noise rates less than 80 for the symmetric case and less than 40 for the asymmetric case. The nature of the final classifier used after the learned representation partially explains these results; we used a single dense layer (see Section \ref{implementationdetails}). This classifier may under-fit as the number of learnable parameters might be too low to actually fit complex datasets such as CIFAR10 and CIFAR100 even with a good given representation. Using more complex classifiers such as Multi-Layer Perceptron could have led to comparable performances than fine tuning even for low noise rates. This point leaves room for further investigation. Having the possibility to fine tune the representation to better fit the classification task induces the risk to actually degrade it.

Outside of well controlled and perfectly clean datasets, practitioners should first consider to learn a self-supervised representation and then either fine tune it or freeze it with classifier learned with robust algorithms. Self-Supervised Learning (SSL) algorithm such as SimCLR seems to perfectly fit this task, but other SSL algorithms could be used and explored.

Another observation from this benchmark is about the difference in performance between all the tested algorithms. Indeed, if we consider part (B) of Table \ref{table-results-cifar10}, for both noise models and all noise rates, the performances between the algorithms are close, around 0.1 point in accuracy with some exceptional data points. It shows that even complex algorithms have a hard time beating simpler approaches when they are compared with an already learned representation.

The same observation can be done for the part (B) of Table \ref{table-results-cifar100} (for CIFAR 100), especially for the symmetric noise. However the differences between algorithms are better put in perspective with this more complex dataset which contains 10 time more classes and 10 time less samples per classes. We notice that some algorithms start to struggle at high symmetric noise rate or for the more complex asymmetric noise model. 
For example, GLC is under-performing against competitors for all cases and is under-performing against its end-to-end version. One reason could be the small size used for the validation dataset as the transition matrix is evaluated on it in a supervised manner. The small number of samples may impact the performance of the transition matrix estimator. Much less so than the estimator proposed by F-Correction which seems to perform fine even on CIFAR100 for all symmetric noises, yet only above average on asymmetric noises. Seeing F-Correction and GLC not performing well on asymmetric noise for both dataset is surprising as these algorithms were both particularly designed for this case.

Lastly we observe on both Tables \ref{table-results-cifar10} and \ref{table-results-cifar100} that algorithms with additional knowledge on the noise model (see Table \ref{knowledge}) have an edge over algorithms that do not, especially on the hardest cases with more classes, higher noise ratio or more complex noise model. CoL requires the noise ratio as its efficiency relies on the hyper parameters value corresponding to the injection of pseudo labels and confidence in model prediction that are dependent of the noise ratio. CoL emerges among the most well rounded and most efficient algorithm for all noise models, noise rates and datasets thanks partially to this additional knowledge. On the other hand, GLC, DIW and MWNet require an additional clean validation dataset in order to estimate the noise model or a proxy of it to correct the learning procedure on the noisy dataset. We could expect these algorithms to perform better than CoL as they would be able to deal with more complex noise models and have a fine-grained policy for correcting noisy samples. Still these algorithms are not able in these experiments to get a better accuracy than CoL and perform on par with it.

Finally we need to emphasize that only two datasets have been used in this study, specially two datasets about image classification. In order to stronger our claims, more experiments should be conducted.

\section{Conclusion}

In this paper our contribution was to suggest new insights about \textit{decoupling} against \textit{end-to-end} deep learning architectures to learn, preserve or promote a good representation in case of label noise. We presented (i) a new view on a part of the state of the art: the ways to preserve the representation (ii) and an empirical study   which  completes the results and the conclusions of other recent  papers \cite{zheltonozhskii2021contrast,ghosh2021contrastive,zhang2020decoupling}. Experiments conducted draw a comprehensive picture of performances by featuring six methods and nine noise instances of three different kinds (none, symmetric, and asymmetric). Our added value for the empirical study is  the comparison between the "freeze" and the "fine tuning" results.

One conclusion we are able to draw is that designing algorithms that preserve or promote good representation under label noise is not the same as designing algorithms capable of learning from scratch a good representation under label noise. To make end-to-end learning succeed in this setup researchers should take a better approach when designing such algorithms.

Another element that emerged from the experiments was the efficiency of both freeze and fine tuning approaches in comparison to the end-to-end learning approach. Even the most complex algorithms such as DIW when trained in an end-to-end manner are not able to beat simple robust loss as GCE when trained with fine tuning. It questions usual experimental protocols of Robust Learning to Label (RLL) noise papers and questions the recent advances in the field. Evaluating RLL algorithms with pretrained architectures should be the norm as it is easy to do so and the most efficient way for practitioners to train model on noisy data.

One more strong point in this conclusion is that in presence of noise the experiments show that fine tuning of Contrastive representation allows the six methods to achieve better results than their end-to-end learning version and represent a new reference compare to the recent state of art. Results are also remarkable stable versus the noise level.

Since fine-tuned representations are shown to outperform frozen ones, one can conclude that noise-robust classification heads are indeed able to promote meaningful representations if provided with a suitable starting point (contrastingly to readers of \cite{zheltonozhskii2021contrast,ghosh2021contrastive} who might prematurely jump to the inverse conclusion).

However these experiments could be extended to be more exhaustive in two ways: (i) SimCLR is not the only recent and efficient contrastive learning algorithms, MOCO \cite{he2019moco,chen2020mocov2} or Bootstrap Your Own Latent (BYOL) \cite{NEURIPS2020_f3ada80d} could have been used as said earlier in the paper, but other self-supervised or unsupervised algorithms could have been used such as Auto-Encoder \cite{Kramer1991} or Flow \cite{flowiee}; 
(ii) experiments could be extended to datasets from other domains such as text classification or time series classification.

\section*{Acknowledgements}

We would like to thank the anonymous reviewers for their careful, valuable and constructive reviews as well as the words of encouragement on our manuscript. 

\bibliographystyle{splncs04}
\bibliography{references.bib}

\end{document}